\documentclass[preprint,12pt,authoryear]{elsarticle}

\usepackage[utf8]{inputenc}
\usepackage{textgreek}
\usepackage{amssymb}
\usepackage{amsmath}
\usepackage{graphicx}
\usepackage{booktabs}
\usepackage{multirow}
\usepackage{tabularx}
\usepackage{siunitx}
\usepackage{hyperref}
\usepackage{xcolor}
\usepackage[autostyle]{csquotes}
\usepackage{float}
\usepackage{placeins}

\usepackage[
left=1in,
right=1in,
top=1in,
bottom=1in
]{geometry}

\biboptions{authoryear}

\journal{Computer-Aided Civil and Infrastructure Engineering}

\begin{document}

\begin{frontmatter}

\title{A Physics-informed Neural Network Approach for Robust Buckling Load Prediction and Reliability-Based Design of Thin Truncated Conical Shells}

\author[a]{Devasmit Dutta}
\author[a]{Budhaditya De}
\author[b]{Rohan Majumder}
\author[c]{Sudip Kumar Mishra}

\affiliation[a]{organization={Department of Civil and Environmental Engineering, University of California Los Angeles},
city={Los Angeles},
state={CA},
country={USA}}

\affiliation[b]{organization={Department of Civil and Infrastructure Engineering, Adani University},
city={Ahmedabad},
state={Gujarat},
country={India}}

\affiliation[c]{organization={Department of Civil Engineering, Indian Institute of Technology Kanpur},
city={Kanpur},
state={Uttar Pradesh},
country={India}}

\begin{abstract}
Thin-walled truncated conical shells are widely used in aerospace, marine, offshore, and lightweight infrastructure systems due to their high strength-to-weight ratio and geometric efficiency. However, their buckling resistance under axial compression is highly sensitive to geometric imperfections, manufacturing tolerances, material variability, and nonlinear instability effects. Conventional shell design procedures therefore rely on conservative knockdown factors (KDFs), such as those recommended in NASA SP-8019, to reduce theoretical elastic buckling loads. Although such deterministic recommendations are useful for ensuring safety, they do not explicitly account for shell geometry, fabrication quality, data uncertainty, or target reliability level. This study develops a physics-informed neural network (PiNN) framework for robust prediction of critical buckling loads of thin truncated conical shells and integrates the trained surrogate within a reliability-based design (RBD) formulation. The proposed model combines original geometric and material descriptors with mechanics-informed features derived from shell stability theory and the localized reduced stiffness method (LRSM). A physics-informed loss function is also introduced to penalize mechanically inadmissible predictions that exceed the theoretical elastic buckling load. The framework is trained and evaluated using 133 experimental Mylar conical shell tests under axial compression. Compared with a conventional deep neural network (DNN), the PiNN achieves improved predictive accuracy, reduced mean absolute error, and enhanced physical consistency. The trained PiNN surrogate is then used to evaluate reliability indices and calibrate safety-consistent KDFs for prescribed target reliability levels. Results show that the proposed PiNN-RBD framework can generate less conservative yet reliability-consistent design KDFs, providing an efficient computational tool for uncertainty-aware design of imperfection-sensitive shell structures.
\end{abstract}


\end{frontmatter}


\section{Introduction}
\label{sec:introduction}

Thin-walled shell structures are among the most efficient load-carrying systems used in modern engineering applications as they offer excellent strength and stiffness at lower structural masses. Truncated conical shells, in particular, are widely employed in aerospace launch vehicles, spacecraft adapters, offshore and marine structures, pressure vessels, and advanced lightweight infrastructure systems. Despite their obvious structural advantages, their design, however, is often governed by stability concerns rather than material strength. Under axial compression, small geometric imperfections, boundary-condition deviations, residual stresses, material variability, and nonlinear mode interactions can reduce the experimental buckling load far below the classical elastic prediction  \citep{koiter1945stability,seide1961buckling,seide1962survey,singer2002buckling, hutchinson2016buckling}.

Whereas yielding or material failure generally depends on stress levels exceeding material strength, shell buckling can occur as a consequence of geometric instability at load levels substantially below the theoretical elastic critical load. Because of this strong imperfection sensitivity, shell design has traditionally relied on empirical knockdown factors (KDFs), such as those recommended in NASA SP-8019 for truncated conical shells  \citep{seide1968buckling}. These factors provide a practical safety margin but are deterministic and often conservative because they do not explicitly reflect shell geometry, fabrication quality, data uncertainty, or target reliability level. High-fidelity nonlinear finite element analysis can model imperfections and nonlinear collapse more directly and has supported recent robust KDF developments \citep{wagner2018robust, wagner2019robust}. However, repeated nonlinear analyses remain computationally expensive for reliability analysis, uncertainty quantification, optimization, and reliability-based calibration of design factors. 

Over the recent past, machine learning (ML) has emerged to be a promising alternative for developing computationally efficient surrogate models capable of approximating complex structural behavior. Data-driven surrogate models offer an efficient alternative for structural stability prediction. Artificial neural networks have been used to predict the buckling loads of axially compressed thin cylindrical shells using experimental databases \citep{rehman2017effect, mandal2021application}. Recent studies have extended machine-learning-based buckling prediction to composite cylindrical shells \citep{guan2023predicting}, image-based imperfection fields and buckling modes \citep{hao2023image}, random-field imperfection inputs for stochastic buckling analysis \citep{schweizer2025artificial}, and KDF prediction for truncated conical shells using ANN-based metamodeling \citep{majumder2024predicting}. Subsequent studies have explored
SVR, ensemble decision-tree and Gaussian-process models \citep{majumder2026robust}, as well as hybrid Gaussian-process/XGBoost frameworks \citep{majumder2026data}. More broadly, machine-learning surrogates have also been explored for structural stability and uncertain buckling analysis under complex parameter variability \citep{shahin2023ann, liu2024hybrid}. These studies demonstrate the computational value of ML; however, purely data-driven approaches exhibit several limitations. Their performance is strongly dependent on the quantity and quality of available data, and they may generate physically unrealistic predictions when extrapolating beyond the range of the training dataset. These limitations are particularly problematic in shell stability problems, where training data are often scarce and the response is governed by physical principles.

Physics-informed machine learning \citep{raissi2019physics,karniadakis2021physics,wang2021understanding} provides a natural route to address this limitation by embedding laws of physics and mechanics into the model through the architecture, input features, governing-equation residuals, boundary constraints, or loss function. Early neural approaches for solving differential equations were introduced by \citep{lagaris1998artificial}, and modern PiNN frameworks were later formalized for forward and inverse problems governed by nonlinear partial differential equations \citep{raissi2019physics}. The broader field of physics-informed machine learning has since been reviewed extensively \citep{karniadakis2021physics,cuomo2022scientific}. In structural mechanics, PiNNs have been applied to shell response prediction using Naghdi shell theory \citep{bastek2023physics} and nonlinear buckling analysis of beams \citep{bazmara2023application} and physics-informed buckling-load prediction of spherical shells \citep{xu2026physics}. These developments suggest that stability-guided learning is particularly suitable for thin-shell buckling, where experimental data are scarce but strong mechanics-based bounds and reduced-stiffness concepts are available.  

This study uses a PiNN framework for robust buckling-load prediction of thin truncated conical shells under axial compression and integrates the trained surrogate into a reliability-based design formulation. The model combines geometric, material, and fabrication-quality descriptors with mechanics-informed features derived from shell stability theory and the localized reduced stiffness method (LRSM). A physics-informed loss term is introduced to penalize predictions that exceed the theoretical elastic buckling load, thereby improving physical admissibility. The framework is evaluated using an experimental database of axially compressed Mylar conical shells and compared with a conventional deep neural network. The trained PiNN surrogate is then used for reliability assessment, PDP-based interpretation of feature influence, and calibration of safety-consistent KDFs for prescribed target reliability indices. The main objective is to provide a computationally efficient and physically interpretable basis for uncertainty-aware design of imperfection-sensitive conical shells.

\section{Physics-informed Machine Learning Model Architecture}
\label{sec:framework}

\subsection{Overview of the proposed framework}

The adopted ML model is developed as a physics-informed surrogate for predicting the experimentally observed axial buckling load of thin truncated conical shells. Unlike a conventional deep neural network (DNN), which learns the mapping between input variables and buckling load solely from data, the present PiNN incorporates shell-stability knowledge at two levels. First, mechanics-informed quantities obtained from shell stability theory and the localized reduced stiffness method (LRSM) (discussed later on) are introduced as additional input features. Second, the loss function is augmented with a physics-based penalty that discourages mechanically inadmissible predictions.

\begin{figure}[htbp]
    \centering
    \includegraphics[width=0.95\textwidth]{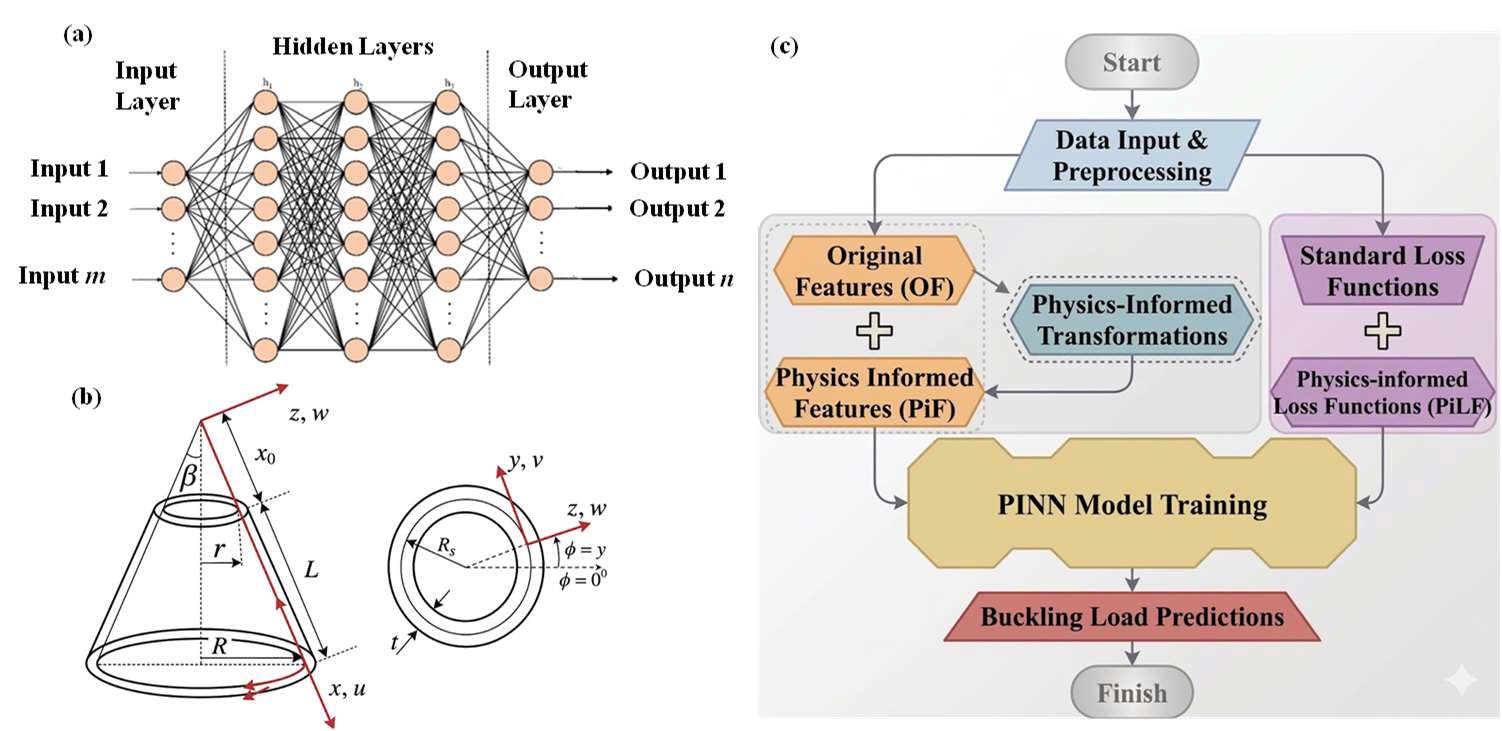}
    \caption{Overview of the proposed physics-informed learning framework: (a) representative deep neural network architecture, (b) truncated conical shell geometry and coordinate definitions, and (c) PiNN workflow combining original features, physics-informed features, and physics-informed loss functions.}
    \label{fig:PiNN_framework}
\end{figure}

The model should, therefore, be interpreted as a physics-guided machine-learning surrogate rather than a classical differential-equation-based PiNN. In the present problem, the objective is not to solve the governing shell equations over the spatial domain. Instead, the objective is to learn a physically consistent regression model that maps shell geometry, material properties, fabrication-quality descriptors, and LRSM-based stability indicators to the experimentally measured critical buckling load. This strategy is particularly suitable for shell buckling problems because experimental datasets are limited, while strong mechanics-based bounds and reduced-stiffness estimates are available from classical and modern shell stability theory. 
Fig.~\ref{fig:PiNN_framework} illustrates the overall PiNN workflow for the present study as well as a typical schematic of truncated conical shells with salient features.

\subsection{Dataset preparation and Feature Engineering }

For each shell specimen, the input vector consisted of original experimental descriptors and physics-informed descriptors. The original variables describe the shell geometry, material behavior, and fabrication quality (relating to surface imperfections), while the physics-informed variables introduce mechanics-based information related to imperfection sensitivity and reduced buckling stiffness. The complete PiNN input vector was defined as

\begin{equation}
\mathbf{x} =
\left[
\beta, L, R, t, E, \nu, \mathrm{FTQC}, \rho_{LRSM}, N_{LRSM}
\right],
\label{eq:input_vector}
\end{equation}
where $\beta$ is the semi-vertex angle, $L$ is the slant length, $R$ is the base radius, $t$ is the shell thickness, $E$ is Young's modulus, $\nu$ is Poisson's ratio, FTQC denotes fabrication tolerance quality control signifying the degree of surface imperfections generated due to manufacturing defects (usually categorized as [1, 2, 3] corresponding to low, moderate and high-degree of surface imperfections), $\rho_{LRSM}$ is the LRSM-based KDF (i.e. ratio of the realistic buckling load to the theoretical counterpart) determined through the localized reduced stiffness method (LRSM) which is discussed in the subsequent sections, and, $N_{LRSM}$ is the LRSM-based buckling load. The first seven variables are treated as original features, whereas the last two are treated as physics-informed features. The performance of the PiNN model was compared with a baseline Deep Neural Network (DNN) model to judge its efficacy.

\subsubsection{Physics-informed Feature Construction}

The localized reduced stiffness method (LRSM), introduced by \citep{wagner2019robust}, is a lower-bound shell-buckling approach developed from the reduced stiffness method (RSM). Its central idea is that, during imperfection-sensitive buckling, local snap-through or dimple formation can strongly reduce the stabilizing membrane stress contribution in a limited region of the shell. Instead of reducing membrane stiffness globally, as in the classical RSM, LRSM reduces the membrane stiffness locally so that the numerical model reproduces the membrane-stress redistribution associated with the plateau or lower-bound buckling load. This localized reduction also avoids prescribing the first buckling eigenmode as an initial imperfection and provides practical lower-bound estimates for cylindrical, conical, and spherical shells under different loading conditions. In the present study, this mechanics-based concept was used as a source of physics-informed features for the PiNN. Specifically, the LRSM-based KDF, ${\rho_{LRSM}}$  and the corresponding LRSM buckling load $N_{LRSM}$ were computed from the shell geometry through the Batdorf parameter Z.
\begin{equation}
Z = \frac{L^2 (1 - \nu^2)}{R_a t},
\label{eq:Z}
\end{equation}
Where $R_a$ is the average radius of the conical shell and $\nu$ is the Poisson’s ratio. For obtaining $\rho_{LRSM}$ as a function of Z, a power law regression was fitted with the KDFs obtained from 290 LRSM-based nonlinear finite element simulations in ABACUS/CAE 2022. For each shell geometry, a localized region of reduced membrane stiffness was introduced into the shell, and geometrically nonlinear buckling analyses were carried out. This localized stiffness reduction was intended to reproduce the membrane-stress redistribution and lower-bound buckling behavior associated with local snap-through instability \citep{wagner2019robust}. The width and/or position of the reduced-stiffness region was varied, and the minimum local buckling load obtained from the LRSM analyses was taken as $N_{LRSM}$, consistent with the design-load definition proposed by \citep{wagner2019robust}. The corresponding LRSM KDF was then computed as

\begin{equation}
\rho_{LRSM} = \frac{N_{LRSM}}{N_{elastic}}
\label{eq:rholrsm}
\end{equation}
where $N_{elastic}$ is the theoretical elastic buckling load for the same shell geometry obtained using the classical equation,

\begin{equation}
N_{elastic}=
\frac{2 \pi E t^2 cos^2(\beta)}{\sqrt{3(1 - \nu^2)}}
\label{eq:nelastic}
\end{equation}

\begin{figure}[ht]
    \centering
    \includegraphics[width=0.8\linewidth]{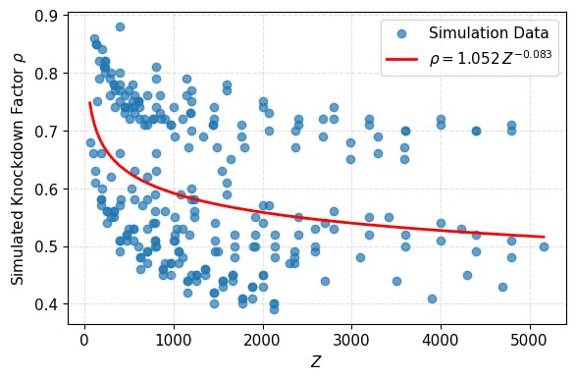}
    \caption{Power law regression fit between simulated $\rho$ using LRSM and Z}
    \label{fig:power law rho fit}
\end{figure}

After repeating this procedure for different shell geometries, the resulting pairs $(Z_i,\rho_{LRSM,i})$ were used to fit a power-law relation of the form $\rho_{LRSM}=aZ^b$. The adopted LRSM regression for this study is shown below in Eq.~\ref{eq:rho_lrsm} while Fig.~\ref{fig:power law rho fit} illustrates it graphically.

\begin{equation}
\rho_{LRSM}= 1.0524\,Z^{-0.0834}
\label{eq:rho_lrsm}
\end{equation}

The associated LRSM-based buckling load was then computed from the regression fit results using Eq.~\ref{eq:rholrsm} and these two quantities were appended to the original features to generate the overall feature set for the PiNN model.

\subsection{Physics-informed Loss Function}

The training objective combines a conventional data-driven regression loss with a physics-informed constraint loss. The total loss is defined as

\begin{equation}
\mathcal{L}_{\text{total}} = \mathcal{L}_{\text{data}} + \lambda \mathcal{L}_{\text{phys}}
\label{eq:total_loss}
\end{equation}

where $\mathcal{L}_{data}$  measures the discrepancy between predicted and experimental buckling loads, $\mathcal{L}_{phys}$  penalizes violation of the elastic buckling bound, and $\lambda$ controls the relative contribution of the physics-informed penalty. In the present study, $\lambda=1$ was adopted. The data loss is expressed as the mean squared error,

\begin{equation}
\mathcal{L}_{\text{data}} = \frac{1}{n}\sum_{i=1}^{n}\left(N_{\text{pred},i} - N_{\text{exp},i}\right)^2
\label{eq:data_loss}
\end{equation}

where n is the number of training samples, $N_{pred,i}$ is the PiNN-predicted buckling load, and $N_{exp,i}$ is the experimentally observed buckling load. The physics-informed loss is defined as,

\begin{equation}
\mathcal{L}_{\text{phys}} = \frac{1}{n}\sum_{i=1}^{n}\left[\text{ReLU}\left(N_{\text{pred},i} - N_{\text{elastic},i}\right)\right]^2
\label{eq:phys_loss}
\end{equation}

This loss term is activated only when the predicted buckling load exceeds the corresponding theoretical elastic buckling load. Therefore, the penalty is one-sided: predictions below the elastic buckling load are not penalized by $L_{phys}$, whereas predictions above the elastic limit receive a quadratic penalty. This is physically meaningful because geometric imperfections, boundary deviations, residual stresses, and nonlinear instability effects generally reduce the experimentally observed buckling resistance relative to the ideal elastic value. The ReLU-based formulation has an important practical advantage. It does not force the network prediction to follow the elastic solution directly; rather, it only discourages predictions that violate a known upper bound. The PiNN can therefore learn the experimentally observed reduction in buckling resistance while avoiding unconservative predictions that exceed the ideal-shell capacity.

\subsection{PiNN Model Selection and Training}
The experimental database consisted of 133 Mylar truncated conical shell tests \citep{seide1962survey} subjected to axial compression. The dataset included shells with different geometric configurations and fabrication tolerance levels. Table~\ref{tab:features} shows the entire feature set including original and physics-informed features along with their ranges.

\begin{table}[htbp]
\centering
\caption{Original and physics-informed features used in the PiNN framework.}
\label{tab:features}
\begin{tabular}{lll}
\toprule
\textbf{Feature type} & \textbf{Feature} & \textbf{Range / description} \\
\midrule
Original & Semi-vertex angle, $\beta$ & $10^\circ$--$60^\circ$ \\
Original & Slant length, $L$ & 35.52--366.77 mm \\
Original & Base radius, $R$ & 99.73--169.17 mm \\
Original & Thickness, $t$ & 0.05--0.25 mm \\
Original & Young's modulus, $E$ & 4826 MPa \\
Original & Poisson's ratio, $\nu$ & 0.30--0.33 \\
Original & FTQC & 1, 2, 3 \\
Physics-informed & LRSM KDF, $\rho_{LRSM}$ & Derived feature \\
Physics-informed & LRSM load, $N_{LRSM}$ & Derived feature \\
\bottomrule
\end{tabular}
\end{table}

The complete dataset was divided into training and testing subsets using an $80:20$ split. The training data were used to optimize the network parameters, while the testing data were reserved for evaluating generalization performance. The model parameters were optimized using the Adam optimizer \citep{kingma2015adam} with a fixed learning rate ($1e^{-3}$) over 500 epochs, ensuring convergence of the combined data-driven and physics-informed losses. A minibatch training strategy was adopted which offered improved generalization and computational efficiency. Furthermore, a systematic non-uniform architecture search involving varied number of hidden layers as well as the number of neurons per layer was conducted to identify the optimal model configuration for enhanced predictive accuracy, consistency and stability. 

\section{Results and Discussion}

\begin{figure}[H]
    \centering
    \includegraphics[width=\linewidth]{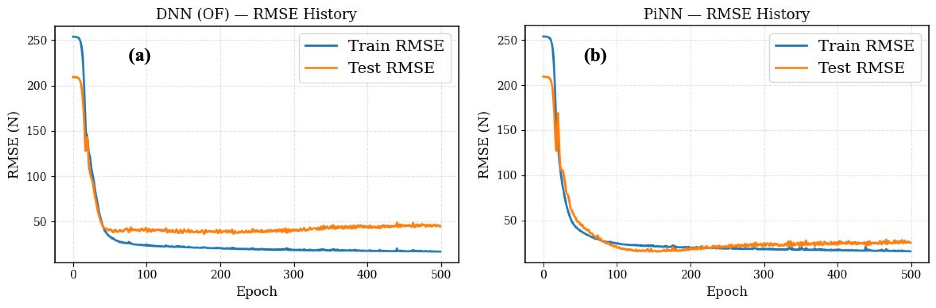}
    \caption{RMSE model loss history as function of epochs for baseline DNN and proposed PiNN models}
    \label{rmse_log}
\end{figure}

\begin{figure}[H]
    \centering
    \includegraphics[width=\linewidth]{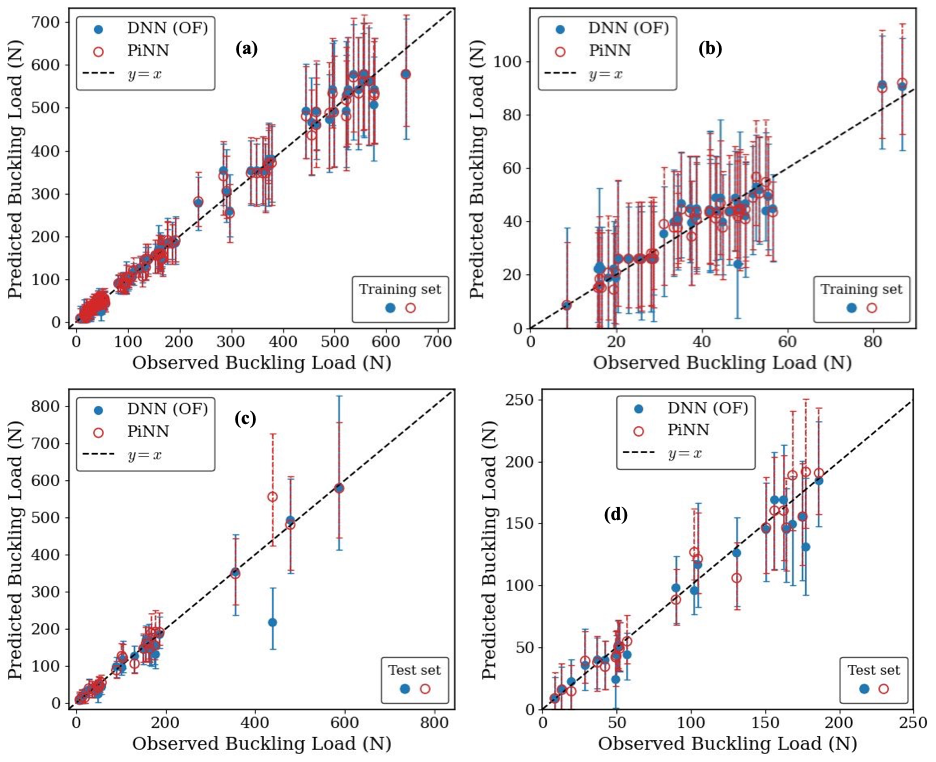}
    \caption{Buckling load predictions $N_{pred}$ against the experimental values for the PiNN model and traditional DNN for training and test datasets. Plots (b) and (d) show enlarged views of plots (a) and (c) within the 0-90 N and 0-250 N range.}
    \label{fig:npred_vs_obs}
\end{figure}

The predictive performance of the PiNN was evaluated using the coefficient of determination ($R^2$), root mean square error (RMSE), mean absolute error (MAE), and empirical 95\% confidence interval (CI) coverage. The root mean squared error (RMSE) model loss as a function of the number of epochs is presented in Fig.~\ref{rmse_log} for illustration. The conventional DNN shows a widening gap between train and test RMSE after convergence, indicating slight overfitting and weaker generalization. In contrast, the PiNN maintains closely aligned train and test RMSE curves throughout training, demonstrating more stable learning and better generalization performance. Fig.~\ref{fig:npred_vs_obs} shows the buckling load predictions ($N_{pred}$) with 95\% confidence interval (CI) bands against the experimental values for the PiNN model along with those from a traditional baseline DNN for both training and test datasets while Table~\ref{tab:perf_metric} tabulates the different performance metrics for both the PiNN and DNN models. 

\begin{table}[H]
\centering
\caption{Comparison of predictive performance between the baseline DNN and the proposed PiNN for the training and test datasets.}
\label{tab:model_performance}
\renewcommand{\arraystretch}{1.2}
\begin{tabular}{lcccc}
\toprule
\multicolumn{5}{c}{\textbf{Training Set}} \\
\midrule
\textbf{Model} & $\mathbf{R^2}$ & \textbf{RMSE (N)} & \textbf{MAE (N)} & \textbf{95\% CI Coverage} \\
\midrule
DNN  & 0.9916 & 16.81 & 9.64 & 0.991 \\
PiNN & 0.9925 & 15.74 & 9.32 & 0.996 \\
\midrule
\multicolumn{5}{c}{\textbf{Test Set}} \\
\midrule
\textbf{Model} & $\mathbf{R^2}$ & \textbf{RMSE (N)} & \textbf{MAE (N)} & \textbf{95\% CI Coverage} \\
\midrule
DNN  & 0.906 & 44.76 & 17.47 & 0.926 \\
PiNN & 0.971 & 25.06 & 12.46 & 0.963 \\
\bottomrule
\end{tabular}
\label{tab:perf_metric}
\end{table}

\begin{figure}
    \centering
    \includegraphics[width=\linewidth]{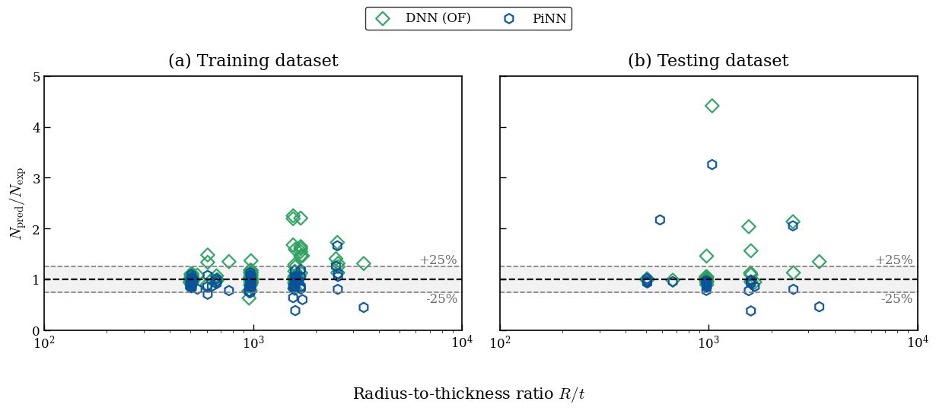}
    \caption{$N_{pred}/N_{exp}$ for the baseline DNN and PiNN models over the radius-to-thickness ratio R/t, with the dashed lines indicating the ±25\% engineering acceptance band}
    \label{fig:25_percent}
\end{figure}

The predictive performance of the proposed PiNN model was compared with that of the conventional DNN using both training and testing datasets. In the training phase, the PiNN showed slightly better performance than the DNN, achieving $R^2=0.9925$, RMSE =15.74N, and MAE =9.32N, compared with the DNN values of $R^2=0.9916$, RMSE =16.81N, and MAE =9.64N. Although the difference in training accuracy is modest, the improvement becomes more significant on the test dataset. The conventional DNN exhibited a noticeable reduction in generalization performance, with test $R^2=0.906$, RMSE =44.76N, and MAE =17.47N. In contrast, the PiNN achieved a substantially higher test $R^2=0.971$, while reducing the test RMSE to 25.06 N and the test MAE to 12.46 N. Relative to the DNN, the PiNN reduced the test RMSE by approximately 44.0\% and the test MAE by approximately 28.7\%, while increasing the test $R^2$ from 0.906 to 0.971. The PiNN also improved the empirical 95\% confidence interval coverage from 0.926 to 0.963, indicating better robustness against prediction uncertainty. Overall, these results demonstrate that incorporating LRSM-derived physics-informed features and the physics-informed loss constraint improves both predictive accuracy and generalization, enabling the PiNN to produce more mechanically consistent buckling-load predictions than the purely data-driven DNN. To further highlight the efficacy of the proposed PiNN model, Fig.~\ref{fig:25_percent} compares the normalized prediction ratio, $N_{pred}/N_{exp}$, for the DNN and PiNN models over the radius-to-thickness ratio R/t, with the dashed lines indicating the ±25\% engineering acceptance band. 

In the training dataset, both models clustered around the ideal value of unity, but the DNN showed greater scatter and more pronounced over-predictions. This difference became more evident in the testing dataset, where the DNN exhibited several large outliers, including severe unconservative over-predictions. The PiNN predictions were more tightly concentrated near $N_{pred}/N_{exp}=1$, indicating improved consistency across different shell slenderness levels. This trend is consistent with the higher acceptance coverage of the PiNN, which achieved 96.2\% and 88.9\% coverage in the training and testing datasets, respectively, compared with 94.3\% and 85.2\% for the DNN. Overall, these results reinforce the fact that the incorporation of LRSM-derived physics-informed features and the physics-based loss constraint improved not only the average predictive accuracy, but also the engineering reliability of individual buckling-load predictions, which is essential for subsequent reliability-based design. The proposed PiNN framework, therefore, provides a more robust and physically consistent surrogate for buckling-load prediction of imperfection-sensitive conical shells than the traditional DNN.

\subsection{Feature Sensitivity Analysis}
\label{sec:training}

\begin{figure}[H]
    \centering
    \includegraphics[width=0.8\linewidth]{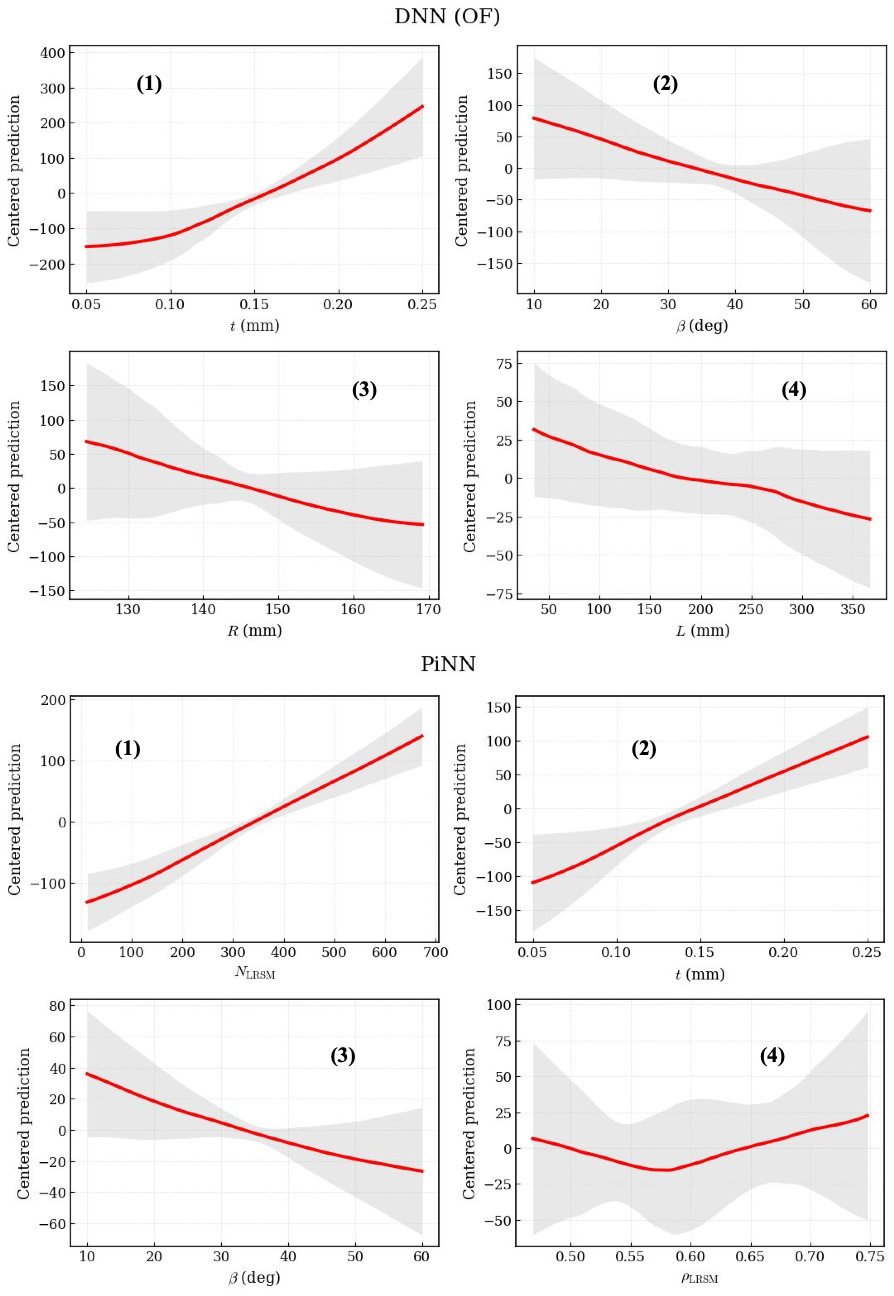}
    \caption{PDPs for the four most influential features affecting buckling strength prediction of thin conical shells using baseline DNN and proposed PiNN models.}
    \label{pdp}
\end{figure}

A feature sensitivity study was conducted using partial dependence plots (PDP). These plots represent the marginal effect of each input feature on the model-predicted buckling load while averaging out the influence of the other variables. Fig.~\ref{pdp} shows the PDPs for the DNN and the PiNN models with the four most influential features affecting the buckling load prediction. The y-axis represents the centered prediction, i.e. the change in predicted buckling load relative to the model’s average prediction. The results reveal that the physics-informed feature $N_{LRSM}$ is the most dominant contributor to model predictions, exhibiting significantly higher importance than all other inputs. Among the geometric parameters, the shell thickness t, semi-vertex angle $\beta$ and slant length L show notable influence. Physics-informed features derived from analytical stability relations considerably influence predictions, confirming that the model captures physically meaningful behavior consistent with established shell buckling mechanics. 

The PDP results reveal that the conventional DNN and the proposed PiNN rely on noticeably different feature-response relationships when predicting the buckling load. For the baseline DNN model, the most influential variables are the original geometric parameters t, $\beta$, R, and L. The predicted buckling load increases strongly with shell thickness t, which is physically reasonable because thicker shells have greater bending and membrane stiffness. However, the DNN also predicts a decrease in buckling resistance with increasing semi-vertex angle $\beta$, radius R, and slant length L. While the decreasing trend with L and $\beta$ can be physically justified in terms of increased slenderness and reduced axial stiffness contribution, the wide confidence bands indicate that the DNN response is more uncertain, especially near the boundaries of the data range. This suggests that the DNN learned broad empirical correlations from the original features, but with relatively high sensitivity to data sparsity and feature interactions.

  For the PiNN model, the most important sensitivity is associated with the LRSM-based buckling load $N_{LRSM}$, which shows a strong and nearly monotonic positive relationship with the centered prediction. This is physically meaningful because $N_{LRSM}$ represents a mechanics-informed lower-bound estimate of the shell buckling capacity; therefore, an increase in $N_{LRSM}$ should directly increase the predicted buckling load. The PDP for $\rho_{LRSM}$ shows a weaker and mild non-monotonic trend compared with $N_{LRSM}$. This is expected because $\rho_{LRSM}$ is a normalized KDF, whereas $N_{LRSM}$  already combines the KDF with the elastic buckling load, which had a very large, scattered range of values. Thus, $N_{LRSM}$ carries more direct information about the absolute buckling resistance. The relatively broader uncertainty band for $\rho_{LRSM}$ also suggests that its marginal effect depends on interactions with other variables such as t, L, $R_a$, and $\beta$. A key implication of these results is that the PiNN does not rely solely on raw geometric correlations. Instead, it learns a response surface guided by physically meaningful quantities, particularly $N_{LRSM}$, while still retaining sensitivity to important original variables such as t and $\beta$. The comparison also supports the role of physics-informed feature construction: the DNN infers buckling behavior indirectly from t, R, L, and $\beta$, whereas the PiNN is provided with LRSM-based descriptors that already encode reduced-stiffness and imperfection-sensitive buckling behavior. This helps explain why the PiNN achieved better test-set accuracy and improved generalization compared with the conventional DNN.

\subsection{Reliability-based Design (RBD)}

The relationship between the selected design variable, i.e., the semi-vertex angle ($\beta$) in this study, and the predicted critical buckling load ($N_{\mathrm{pred}}$) was represented using a metamodel. The metamodel was obtained by fitting the PiNN-predicted buckling loads as a function of the semi-vertex angle using an expression of the following form:

\begin{equation}
N_{\mathrm{pred}}(\beta)
=
3.04\times10^{-3}\beta^{3}
+
9.64\times10^{-2}\beta^{2}
-
14.9\,\beta
+
273.37
\label{eq:metamodel}
\end{equation}

The corresponding PiNN-based KDF was then evaluated by normalizing the predicted critical load with respect to the theoretical elastic buckling load, as

\begin{equation}
\rho_{\mathrm{PiNN}}(\beta)
=
\frac{N_{\mathrm{pred}}(\beta)}
{N_{\mathrm{elastic}}(\beta)}
\label{eq:rho_PiNN}
\end{equation}

Here, $N_{\mathrm{elastic}}$ is the classical elastic critical buckling load of the corresponding perfect conical shell obtained from Eq.~\eqref{eq:nelastic}. Thus, $\rho_{\mathrm{PiNN}}$ represents the actual resistance predicted by the trained PiNN model in normalized form. The limit-state function (LSF) is then defined in terms of $\rho_{\mathrm{PiNN}}$ and the deterministic design $\rho$ as

\begin{equation}
g\!\left(\rho_{\mathrm{PiNN}},\bar{\rho}\right)
=
\rho_{\mathrm{PiNN}}(\beta)
-
\bar{\rho}
\label{eq:lsf}
\end{equation}

where $\bar{\rho}$ is the limiting K adopted from code-based recommendations. In this study, the limiting $\rho$ was taken from NASA SP-8019 \citep{seide1968buckling} as
$\bar{\rho}=0.33$.
The safe and unsafe domains are therefore expressed as,

\begin{equation}
g\!\left(\rho_{\mathrm{PiNN}},\bar{\rho}\right)
=
\begin{cases}
\ge 0, & \text{safe},\\
<0, & \text{failure}.
\end{cases}
\label{eq:safe}
\end{equation}

\begin{figure}
    \centering
    \includegraphics[width=0.5\linewidth]{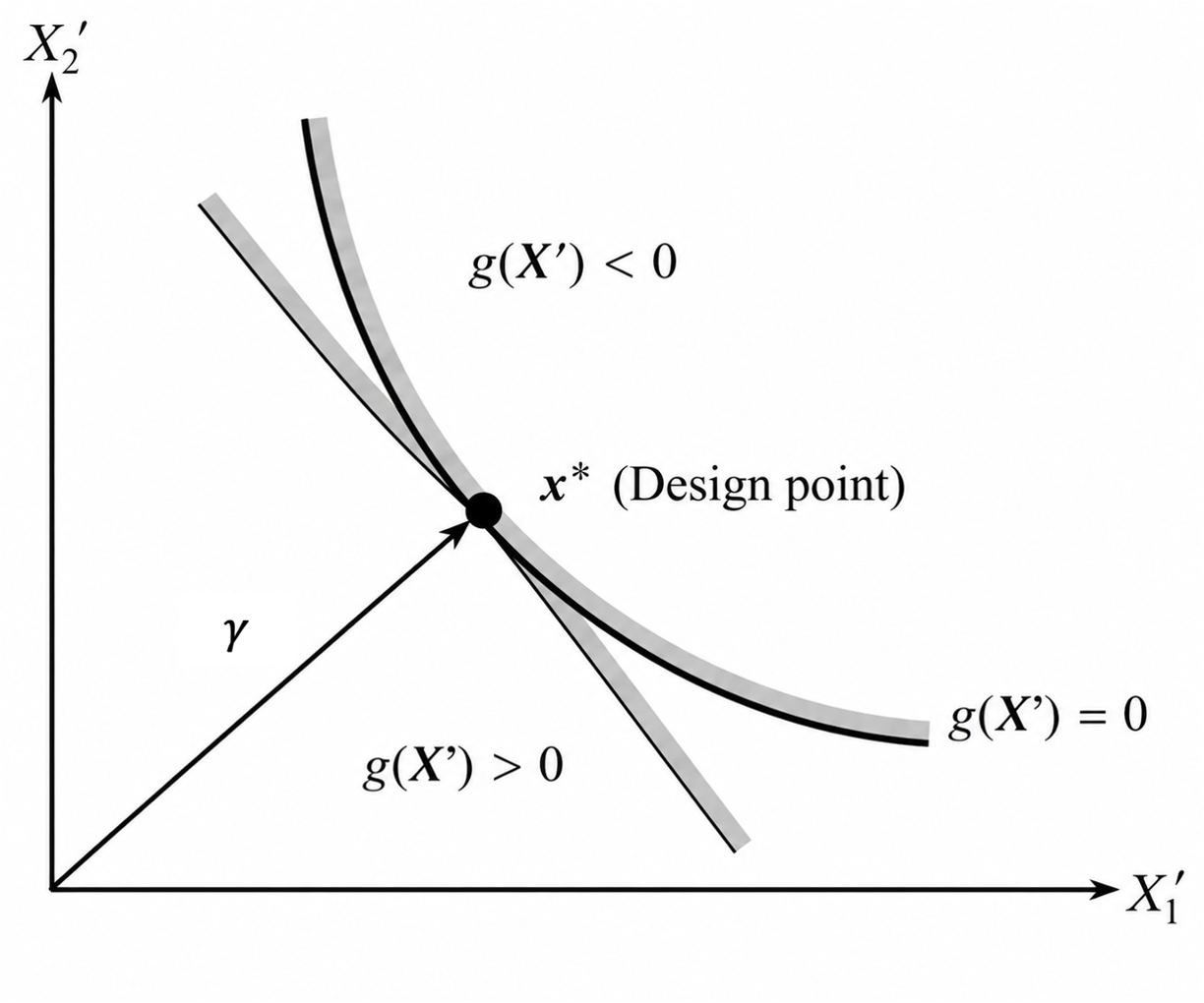}
    \caption{Geometric interpretation of the reliability index $\gamma$ and the design point in standard normal space.}
    \label{fig:8}
\end{figure}

Thus, failure occurs when $\rho_{\mathrm{PiNN}}$ falls below the deterministic design threshold $\bar{\rho}$. Following standard structural reliability theory \citep{melchers1999structural,derkiureghian2005first}, the reliability index $\gamma$ is defined as the minimum distance from the origin in the reduced coordinate system (standard normal space, characterized by zero mean and unit standard deviation) to the limit-state surface (LSF). The point on the LSF corresponding to this minimum distance is referred to as the \textit{design point}, as schematically illustrated in Fig.~\ref{fig:8}. For nonlinear and complex LSFs, determination of the reliability index $\gamma$ and the associated design point can be formulated as a constrained optimization problem, expressed as,

\begin{equation}
\gamma
=
\min_{\mathbf{u}}
\sqrt{\mathbf{u}^{T}\mathbf{u}}
\quad
\Big|
\quad
g\!\left(\rho_{\mathrm{PiNN}},\bar{\rho}\right)
=
0
\label{eq:form}
\end{equation}

Herein, $\mathbf{u}$ is the normally transformed feature random variable vector and solution $\mathbf{u}^{*}$ is the most probable failure point, i.e., the design point. For nonlinear LSFs, this constrained minimization problem is usually solved using iterative first-order reliability method (FORM), such as the Rackwitz--Fiessler algorithm \citep{rackwitz1978structural}, which was also implemented in this study. Once $\gamma$ is obtained, the probability of failure is computed as,

\begin{equation}
P_f = \Phi(-\gamma)
\label{eq:pf}
\end{equation}

where, $\Phi(\cdot)$ is the standard normal cumulative distribution function. The forward reliability analysis evaluates the reliability level associated with a prescribed design variable $\beta_d$. For shell design purposes, however, the inverse problem is often more useful, i.e., determining the design variables (and in turn the design KDFs) that achieve a specified target reliability index $\gamma_T$. The reliability-consistent $\beta$ is obtained by first solving for an optimal value of the design variable (in our case, the semi-vertex angle $\beta$) through an outer optimization workflow which, in the present study, was achieved through the particle swarm optimization algorithm \citep{kennedy1995particle}. Thus, the design variable optimization problem can be written as,

\begin{equation}
\hat{\beta}_d
=
\arg\min_{\beta_d}
\left|
\gamma(\beta_d)-\gamma_T
\right|
\label{eq:beta_opt}
\end{equation}

So, starting off with a specified $\gamma_T$ and trial guess value of the design variable, the forward reliability analysis is carried out which gives us the most probable design variable value and reliability index. The optimization algorithm is continued until the obtained $\gamma$ closely matches $\gamma_T$. Once the error is within acceptable tolerance, the design variable optimum is substituted into the metamodel (i.e., Eq.~\eqref{eq:metamodel}) and the safety-consistent $\rho$'s are obtained.

\begin{figure}[H]
    \centering
    \includegraphics[width=0.8\linewidth]{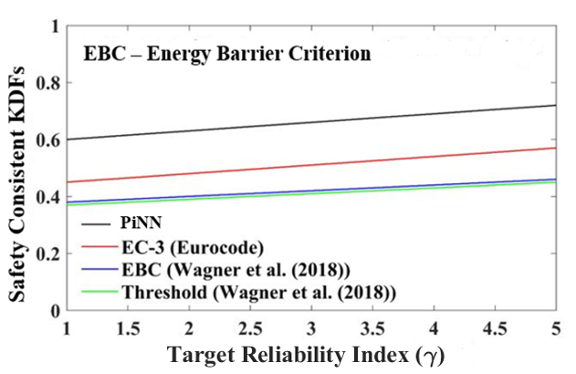}
    \caption{Contrasting PiNN-based and code/literature-based safety-consistent KDFs for varying target reliability indices ($\gamma_T$) for $\frac{R_a}{t}=480$,$\frac{L}{R_a}=1$ and 15\% CoVs in $R_a$ and t.}
    \label{fig:kdf_tri}
\end{figure}

In this study, target reliability indices in the range
\[
1 \le \gamma_T \le 5
\]
were considered which are representative of typical engineering reliability requirements. Fig.~\ref{fig:kdf_tri} presents the safety-consistent KDFs obtained from the reliability-based design framework for different target reliability indices, considering 15\% coefficients of variation (CoV) in average radius ($R_a$) and thickness ($t$). The PiNN-based KDF curve remains consistently higher and increases from approximately 0.60 to 0.72 whereas the Eurocode/ECCS design recommendations \citep{eccs2008buckling}, Energy Barrier Criterion (EBC) \citep{wagner2018robust}, and threshold \citep{wagner2018robust} based curves remain noticeably lower over the full range of target reliability indices, indicating that the proposed framework predicts less conservative yet reliability-consistent design factors. This suggests that the PiNN-RBD framework can recover additional usable buckling capacity while still maintaining the prescribed reliability level.

\section{Conclusions and Future Scope of Research}
\label{sec:conclusion}

This study developed a physics-informed neural network framework for robust buckling-load prediction and reliability-based design of thin truncated conical shells under axial compression. The proposed PiNN was formulated as a physics-guided surrogate model rather than a conventional differential-equation-based PiNN. Shell-stability knowledge was incorporated in two ways: first, through LRSM-derived physics-informed features, namely the LRSM KDF, $\rho_{LRSM}$ and LRSM buckling load $N_{LRSM}$ and second, through a one-sided physics-informed loss function that penalized predicted buckling loads exceeding the theoretical elastic buckling load. This allowed the model to learn from experimental observations while remaining constrained by physically admissible stability behavior.

The framework was trained and evaluated using an experimental database of 133 axially compressed Mylar truncated conical shells and compared with a conventional baseline DNN trained using only original geometric, material, and imperfection-quality descriptors. The PiNN showed slightly improved training performance but substantially better testing performance. On the test dataset, the PiNN achieved 97.1\% $R^2$, RMSE = 25.06 N, MAE =12.46 N, and 95\% confidence interval coverage of 0.963. In comparison, the conventional DNN achieved 90.6\% $R^2$, RMSE = 44.76 N, MAE = 17.47 N, and coverage of 0.926. Thus, the PiNN reduced the test RMSE by approximately 44.0\% and the test MAE by approximately 28.7\%, demonstrating improved generalization and reduced prediction uncertainty. Engineering validation based on the normalized prediction ratio further showed that the PiNN predictions were more tightly clustered around the ideal value and exhibited fewer large unconservative deviations than the DNN.

The feature sensitivity study using partial dependence plots confirmed that the PiNN learned a more physically interpretable response structure. The LRSM-based buckling load emerged as the most influential feature, showing a strong positive relationship with predicted buckling capacity. The model also retained meaningful sensitivity to shell thickness and semi-vertex angle. These results indicate that the PiNN did not rely solely on empirical correlations, but instead used mechanics-informed descriptors to guide the prediction of imperfection-sensitive buckling behavior.

The trained PiNN was further embedded into a reliability-based design framework to calibrate safety-consistent KDFs for prescribed target reliability indices. The resulting PiNN-based KDFs were consistently higher than conventional code and literature-based recommendations, indicating that the proposed method can recover additional usable buckling capacity while maintaining the desired reliability level.

Despite these promising results, the study is limited by the size and scope of the experimental database, which consists only of Mylar conical shells. Future research should expand the database to include metallic, composite, stiffened, and full-scale shell systems, incorporate additional uncertainty sources such as measured imperfections and boundary-condition variability, and develop stronger physics constraints based on shell governing equations, energy barriers, or nonlinear stability criteria.

\section*{Acknowledgments}

The authors would like to acknowledge the SEED Grant (Ref No.
AU/SRG/2005-01) received from Adani University in carrying out the research work.

\section*{Data Availability Statement}

The data used in this study are based on previously reported experimental shell buckling tests and derived physics-informed features. Processed data and trained model parameters may be made available from the corresponding author upon reasonable request.

\section*{Conflict of Interest}

The authors declare no conflict of interest.

\bibliographystyle{agsm}
\bibliography{references}

\end{document}